\documentclass[journal]{IEEEtran}
\usepackage{amsmath,amsfonts}
\usepackage{makecell}
\usepackage[linesnumbered,ruled,vlined]{algorithm2e}
\usepackage{array}
\usepackage[caption=false,font=normalsize,labelfont=sf,textfont=sf]{subfig}
\usepackage{textcomp}
\usepackage{stfloats}
\usepackage{url}
\usepackage{verbatim}
\usepackage{graphicx}
\usepackage{cite}
\usepackage{xcolor}
\begin{document}

\title{Efficient Online Continual Learning in Sensor-Based Human Activity Recognition}

\author{Yao Zhang, Souza Leite Clayton, and Yu Xiao,~\IEEEmembership{Member,~IEEE,}
        
\thanks{Yao Zhang, Souza Leite Clayton and Yu Xiao are
with the Department of Information and Communications Engineering, Aalto University, Finland. E-mail: \{yao.1.zhang, ext-clayton.souzaleite, yu.xiao\}@aalto.fi}
}

\markboth{Journal of \LaTeX\ Class Files,~Vol.~14, No.~8, August~2021}%
{Shell \MakeLowercase{\textit{et al.}}: A Sample Article Using IEEEtran.cls for IEEE Journals}


\maketitle

\begin{abstract}

Machine learning models for sensor-based human activity recognition (HAR) are expected to adapt post-deployment to recognize new activities and different ways of performing existing ones. To address this need, Online Continual Learning (OCL) mechanisms have been proposed, allowing models to update their knowledge incrementally as new data become available while preserving previously acquired information. However, existing OCL approaches for sensor-based HAR are computationally intensive and require extensive labeled samples to represent new changes. Recently, pre-trained model-based (PTM-based) OCL approaches have shown significant improvements in performance and efficiency for computer vision applications. These methods achieve strong generalization capabilities by pre-training complex models on large datasets, followed by fine-tuning on downstream tasks for continual learning. However, applying PTM-based OCL approaches to sensor-based HAR poses significant challenges due to the inherent heterogeneity of HAR datasets and the scarcity of labeled data in post-deployment scenarios.
This paper introduces PTRN-HAR, the first successful application of PTM-based OCL to sensor-based HAR. Unlike prior PTM-based OCL approaches, PTRN-HAR pre-trains the feature extractor using contrastive loss with a limited amount of data. This extractor is then frozen during the streaming stage. Furthermore, it replaces the conventional dense classification layer with a relation module network. Our design not only significantly reduces the resource consumption required for model training while maintaining high performance, but also improves data efficiency by reducing the amount of labeled data needed for effective continual learning, as demonstrated through experiments on three public datasets, outperforming the state-of-the-art. The code can be found here: https://anonymous.4open.science/r/PTRN-HAR-AF60/
\end{abstract}


\section{Introduction}
\label{sec:intro}

\IEEEPARstart{T}{he} recognition of human activities using wearable sensors such as Inertial Measurement Unit (IMU) encompasses many practical applications in smart homes \cite{ishimaruReadingTrackersWild2017}, healthcare \cite{liDeepNeuralNetwork2016}, and manufacturing \cite{grzeszickDeepNeuralNetwork2017}. HAR is typically achieved using machine learning models trained on sensor data from a predefined set of activity classes collected from a selected group of subjects. After deployment, these models often need to evolve to recognize new activities or adapt to the distribution shift caused by changes in users' activity patterns due to aging, disease, or simply a distinct manner of performing the same activity \cite{jhaContinualLearningSensorbased2021}. Simply retraining the model on new data can lead to catastrophic forgetting \cite{leiteResourceEfficientContinualLearning2022}. 

Continual learning focuses on preserving existing knowledge while continuously acquiring new knowledge over time \cite{delangeContinualLearningSurvey2021}. In practice, one approach is to update models periodically with batches of new data. Alternatively, models can be updated continuously as new data becomes available. The latter approach, known as online continual learning (OCL), is particularly well-suited for sensor-based HAR since data arrives in a continuous stream.

Previous approaches to OCL for sensor-based HAR, such as LAPNet-HAR \cite{adaimiLifelongAdaptiveMachine2022}, Leite et al. \cite{leiteResourceEfficientContinualLearning2022}, require the entire model to be continuously updated based on streaming data. The resultant computational costs pose significant challenges for deploying these solutions on resource-constrained devices. Given that most computational effort in HAR systems is concentrated on training the feature extractor, an intuitive approach is to freeze the feature extractor post-deployment to reduce the computational costs. One example is pre-trained model-based (PTM-based) OCL approaches, which entail pre-training the feature extractor on extensive and diverse datasets to ensure robust feature extraction capabilities, followed by fine-tuning on downstream tasks for continual learning during the streaming stage \cite{zhouContinualLearningPreTrained2024}.

Although PTM-based OCL has demonstrated higher performance and resource efficiency in the fields of computer vision
\cite{zhouRevisitingClassIncrementalLearning2023} and natural language processing \cite{cossu2024continual}, its application to HAR has not yielded positive results \cite{10502819}. The main challenge lies in the inherent heterogeneity of HAR datasets, which include variations in sensor modalities and placements. This diversity makes it difficult to effectively combine HAR datasets, limiting the amount of training data available for PTM-based OCL. Consequently, in PTM-based OCL for HAR, the extracted features may lack the necessary descriptiveness to accurately recognize new classes or handle existing classes with domain shifts in the subsequent dense classification layer.
Furthermore, it is essential to minimize the demand for labeled samples that represent new changes. Unfortunately, the issue of data efficiency has not been adequately addressed in OCL for HAR.

In this paper, our aim is to address the challenges of resource and data efficiency in PTM-based OCL for sensor-based HAR without performance degradation. We propose PTRN-HAR, the first successful application of PTM-based OCL to sensor-based HAR. PTRN-HAR advances the state-of-the-art from the following two perspectives:
\begin{itemize}
    \item First, we propose pre-training the feature extractor using contrastive loss with a limited amount of data, rather than large datasets, to enhance data efficiency. This feature extractor is frozen during the streaming stage, thereby reducing computational costs.
    \item Second, we introduce a relation module network as an alternative to the commonly used dense classification layer. Experiments on three public datasets demonstrate that a relation module network can effectively interpret the features related to new classes or data affected by domain shifts, even with a frozen feature extractor pre-trained using a limited amount of data. 
\end{itemize}

Our method outperforms existing OCL approaches such as ER-ACE \cite{caccia2021new} and GDumb \cite{prabhu2020greedy}, achieving a 16.7–18.9\% improvement in Macro-F1 score. By shifting the computational burden of training the feature extractor to the pre-deployment stage, we significantly reduce the time and memory required for training the relation module during the streaming stage. Furthermore, by randomly forming support and query sets from the replay data to train the relation module, our method makes efficient use of the limited labeled data for new classes, thereby enhancing data efficiency.

PTRN-HAR achieves comparable or superior performance in terms of accuracy and macro-F1 score compared to the existing OCL approaches, while significantly surpassing them in terms of computational cost, memory usage, and data efficiency.

The remainder of this paper is structured as follows. Section \ref{sec:rw} introduces the background and related work, followed by the methodology in Section \ref{sec:method}. Section \ref{sec:settings} describes the experimental setup. Section \ref{sec:results} presents and discusses the results of the experiments before concluding the work in Section \ref{sec:conclusion}.

\begin{figure*}[!t]
    \centering
    \includegraphics[width=1\linewidth]{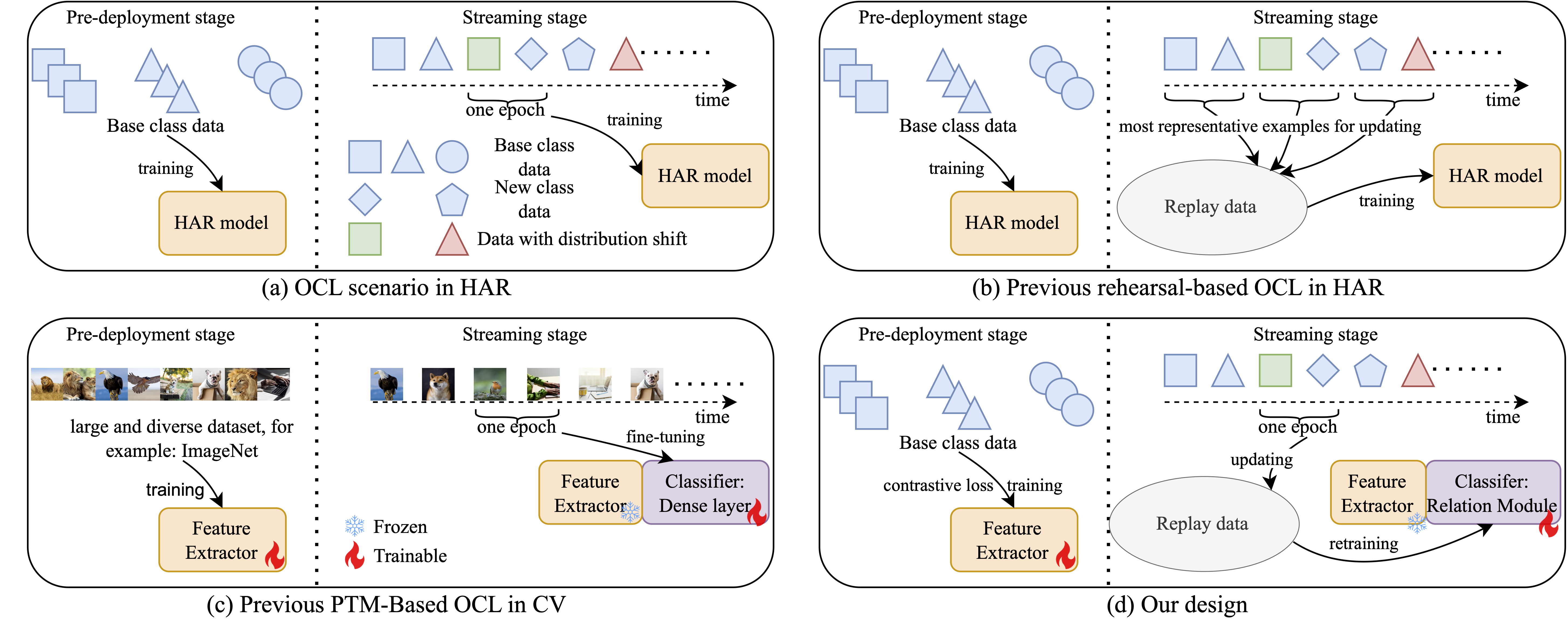}
    \caption{Application scenarios of OCL. Each scenario consists of two stages: pre-deployment and streaming. (a) Traditional OCL for sensor-based HAR typically train the model on base class data during the pre-deployment stage and continuously update the model with streaming data to achieve continual learning. (b) Previous rehearsal-based OCL methods for HAR, such as iCaRL \cite{rebuffiIcarlIncrementalClassifier2017} and OCL-HAR \cite{schiemerOnlineContinualLearning2023}, select representative samples from streaming data to retrain the entire model. (c) PTM-Based OCL methods in the field of computer vision train the feature extractor on large and diverse datasets, and fine-tune the dense layer based on the streaming data. (d) PTNR-HAR extracts more general features through the utilization of contrastive loss. The feature extractor is frozen during the streaming stage, while the relation module is trained on the replay data, permitting higher performance and data efficiency.}
    \label{fig:Comparision of traditional CL and OCL scenarios}
\end{figure*}

\section{Background and Related Work}
\label{sec:rw}

\subsection{Online Continual Learning}
OCL consists of two stages, namely, pre-deployment and streaming. As illustrated in Fig.~\ref{fig:Comparision of traditional CL and OCL scenarios}, a set of base class data \(D_0\) is used to train the initial model \(M_0\) during the pre-deployment stage. After the deployment, the data \(D_t\) arrive in a streaming format without clear task boundaries. The new data comprises both new and old classes, with distributions potentially shifting over time. The goal of OCL is to continuously update the model \(M_0\) with streaming data \(D_t\), to adapt to new classes or distribution shifts that appear in \(D_t\). 

Traditional CL strategies can be categorized into three types: rehearsal-based, regularization-based, and dynamic architecture-based approaches \cite{delangeContinualLearningSurvey2021}. Rehearsal-based methods \cite{rebuffiIcarlIncrementalClassifier2017}, as shown in Fig.~\ref{fig:Comparision of traditional CL and OCL scenarios} (b), achieve CL by storing a subset of data from previous tasks and replaying it regularly as new tasks are learned. Regularization-based methods \cite{liuRotateYourNetworks2018} focus on identifying and preserving the model parameters that are more important for maintaining prior knowledge. Dynamic architecture methods \cite{rusuProgressiveNeuralNetworks2022} expand the model's architecture with the arrival of new tasks, with the newly added components aimed at learning these new tasks while the existing parts remain unchanged to preserve old knowledge. 

Rehearsal-based methods often outperform the other two categories in terms of continual learning performance. However, their reliance on storing replay data imposes memory demands that limit their deployment on edge devices \cite{jhaContinualLearningSensorbased2021, delangeContinualLearningSurvey2021}. Regularization-based methods, while requiring no additional memory, are prone to catastrophic forgetting and tend to exhibit poor generalization performance \cite{li2023fixed}. Dynamic architecture methods achieve stable continual learning without replay data by incrementally expanding the network structure. Nevertheless, as the number of new classes increases, the growing model complexity poses significant challenges for training and deployment on resource-constrained devices \cite{zhang2020regularize}.

\subsection{Related works}

To deploy sensor-based HAR systems on resource-constrained devices, OCL methods for HAR must optimize three key factors: computational cost, memory consumption, and data efficiency. Most existing OCL methods for HAR rely on rehearsal-based approaches, which often struggle to balance these optimization goals while sustaining high performance.

Jha et al. \cite{jhaContinualLearningSensorbased2021} evaluated a range of representative CL methods derived from the computer vision domain, such as iCaRL\cite{rebuffiIcarlIncrementalClassifier2017}, EWC\cite{kirkpatrickOvercomingCatastrophicForgetting2017}, AGEM\cite{chaudhry2018efficient}, on HAR datasets in OCL scenarios. Their findings indicated that rehearsal-based methods outperformed regularization-based and dynamic architecture-based methods in the context of HAR. However, the rehearsal-based methods \cite{rebuffiIcarlIncrementalClassifier2017} require the storage of replay data to update the model, which leads to increased memory consumption. Leite et al. \cite{leiteResourceEfficientContinualLearning2022} considered controlling resource consumption by utilizing Progressive Neural Networks (PNN) \cite{rusuProgressiveNeuralNetworks2022} to reduce network complexity. However, as the number of classes grows, the PNN-based model expands linearly, resulting in longer training and inference durations. Additionally, compressing replay data -- as proposed by the authors to reduce resource consumption -- still leads to some loss of accuracy.

Other methods that do not require a replay memory tend to either be computationally expensive or fail to deliver high performance. For example, Ye et al. \cite{yeContinualActivityRecognition2021} employed a generative adversarial network to generate synthetic sensor data for replay. However, ensuring the quality of the synthetic data poses challenges, and the data generation process causes additional computational costs.

Numerous studies have attempted to tackle the challenges arising from the scarcity of labeled data. For example, LAPNet-HAR \cite{adaimiLifelongAdaptiveMachine2022} addresses the scarcity of labeled data in HAR by combining prototypical networks \cite{snellPrototypicalNetworksFewshot2017} with DeepConvLSTM \cite{ordonezDeepConvolutionalLstm2016}. As illustrated in Fig.~\ref{fig:Comparision of traditional CL and OCL scenarios} (a), the initial HAR model is generated from base class data and is dynamically updated during the streaming stage as new data arrive. However, the performance of LAPNet-HAR varies across different datasets, and updating the model incurs significant computational costs. The experiments of LAPNet-HAR indicated a requisite of at least 200 labeled samples per class to generate sufficiently accurate prototypes.
Similarly, OCL-HAR \cite{schiemerOnlineContinualLearning2023} aimed to minimize the amount of labeled data required for CL in HAR, yet experimentally over 100 labeled samples per class were still needed for OCL-HAR to achieve reasonable performance. Moreover, this method is resource-intensive, as it necessitates retraining the entire model whenever a new class is introduced. In this study, our objective is to achieve OCL with as few as 20 labeled samples per new class.

The emerging PTM-based OCL methods present new opportunities for resource-efficient HAR. Unlike traditional CL methods that rely on replay data or parameter protection, the efficancy of PTM-based CL, as illustrated in Fig.~\ref{fig:Comparision of traditional CL and OCL scenarios} (c), lies in the feature extraction ability of PTM. For instance, SimpleCIL \cite{zhouRevisitingClassIncrementalLearning2023} in the computer vision domain achieves CL by freezing PTM weights and fine-tuning only the final dense layer. However, the approach requires extensive and diverse datasets like ImageNet \cite{5206848} for pre-training. The relatively small sizes of available sensor-based HAR datasets, coupled with the challenge of combining these datasets due to their inherent heterogeneity, necessitate a new design to effectively apply PTM-based OCL approaches to sensor-based HAR. 

\section{Methodology}
\label{sec:method}
In this section, we first provide an overview of PTRN-HAR, then we go into the details of the feature extractor and the relation module.

\begin{figure*}
    \centering
    \includegraphics[width=1\linewidth]{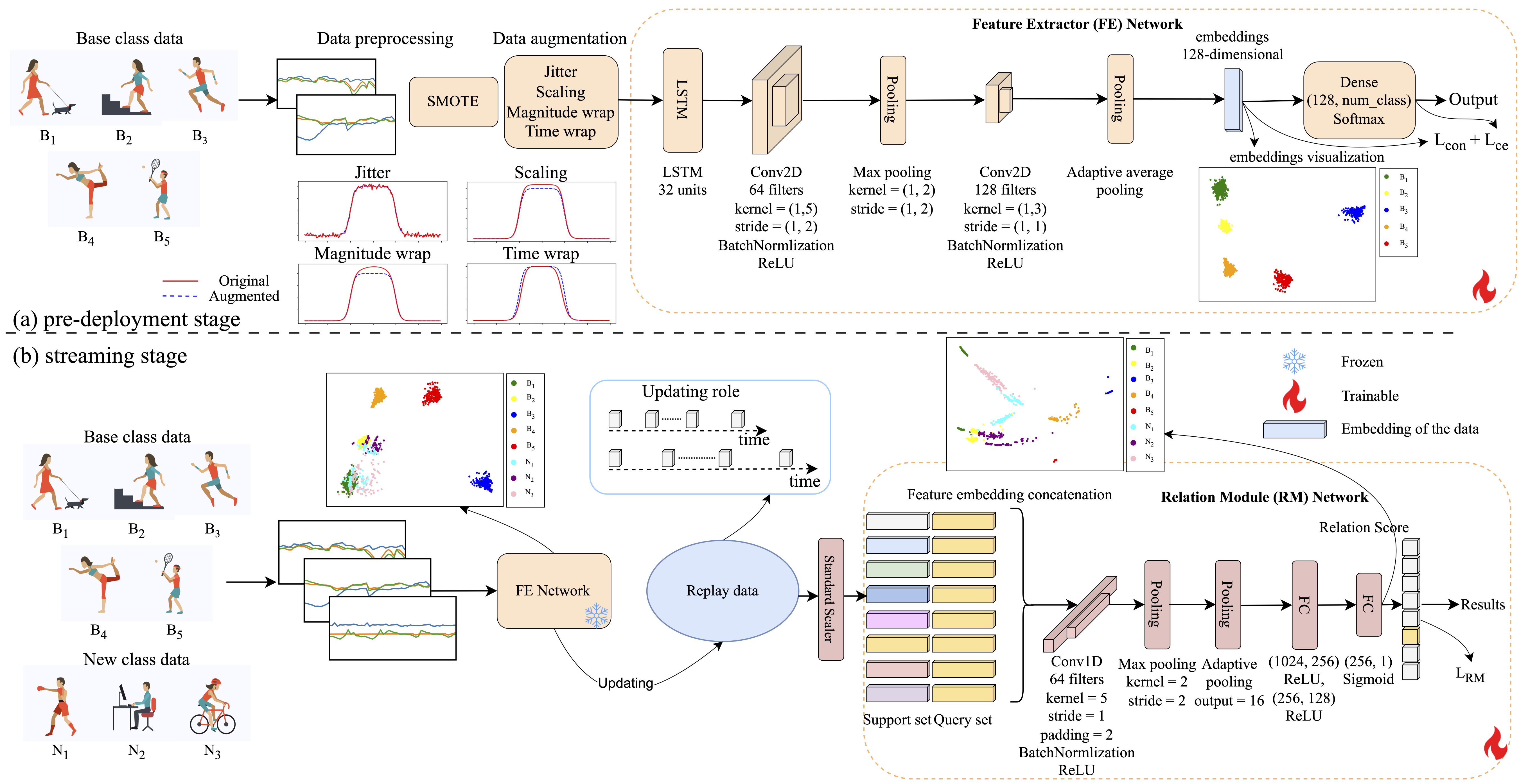}
    \caption{\textbf{Overall pipeline of PTRN-HAR.} (a) In the pre-deployment stage, base class data are used to train the FE network. The embeddings output from the penultimate layer of the FE network are utilized as the features of the data. (b) In the streaming stage, the FE network is frozen, and RM network is used to reclassify the embeddings of streaming data. PTRN-HAR stores \(N\) (default = 20) embeddings for each class as replay data, which is continuously updated based on the incoming labelled streaming data. When a new class emerges or there is a considerably change in the replay data (i.e., domain change), the RM network is retrained using the replay data to enable continual learning. }
    \label{fig:pipeline of PTRN-HAR}
\end{figure*}

\begin{figure*}
    \centering
    \includegraphics[width=1\linewidth]{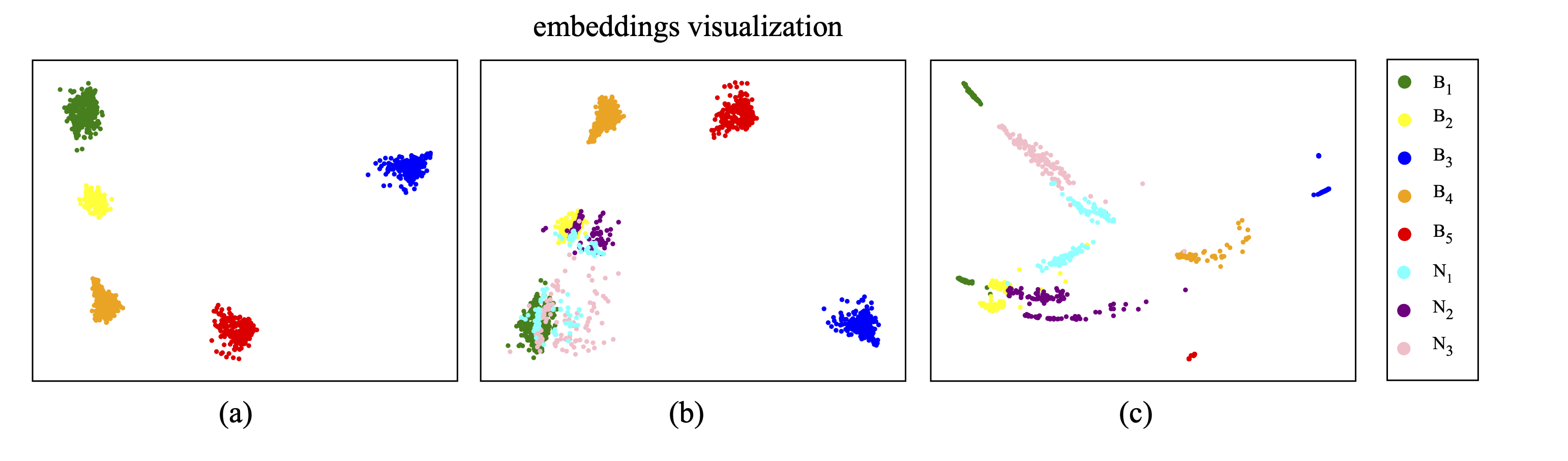}
    \caption{\textbf{The PCA visualization for embeddings after each stage.} Points in different colors represent data from different classes. (B1, B2, B3, B4, B5) denote the base classes, while (N1, N2, N3) represent the new classes. (a) In the pre-deployment stage, the embeddings extracted by the FE network are linearly separable across classes. (b) In the streaming stage, due to the emergence of new classes and the frozen FE network, the embeddings of new and old classes begin to overlap. (c) After reprocessing through the RM network, the embeddings of all classes become linearly separable again. }
    \label{fig:embeddings visualization}
\end{figure*}

\subsection{PTRN-HAR Overview}
\label{sec:PTRN-HAR Overview}

Fig.~\ref{fig:pipeline of PTRN-HAR} presents an overview of PTRN-HAR which consists of two key components: 1) the feature extractor (FE network, Section~\ref{sec:feature extractor}) responsible for extracting general features from the data; and 2) the relation module (RM network, Section~\ref{sec:relation module}) acting as a learnable nonlinear relational classifier to reclassify the features of all classes. The workflow of PTRN-HAR is shown in Algorithm~\ref{algorithm:algorithm 1}. In the pre-deployment stage, the feature extractor is trained on $D_0$, containing the base class set $C_{base}$ and the corresponding data \(\{(x_i, y_i)\}_{i=1}^N\). The output embedding $M$ of the model's penultimate layer is used to represent the general features of the data. We visualize the embeddings using Principal Components Analysis (PCA), as shown in Fig.~\ref{fig:embeddings visualization}(a). The results indicate that the extracted features of base class data by FE network are linearly separable at this stage.

In the streaming stage, where OCL methods are applied, the incoming streaming data contains both new class data $D_1$ and base class data $D_0$. One batch of this streaming data is denoted as \(D_t\), which may contain both unlabelled $D_{tu}$ and labelled data $D_{tl}$. \(D_{tu}\) is used to evaluate the model's performance, while $D_{tl}$ is used to train the model for continual learning. Data $D_{tl}$ are first processed by the FE network \(M_f\) to obtain the corresponding embeddings $E_{tl}$, which are used to update the replay data \(R_t\).(see Scetion \ref{sec:replay update})
As shown in Fig.~\ref{fig:embeddings visualization}(b), the embeddings corresponding to $D_0$ remain linearly separable, while the embeddings of $D_1$ are scattered and partially overlap with $D_0$ since the FE network has not been trained on $D_1$. 
However, these scattered embeddings still retain valuable information, which is why the RM network is employed to reclassify them. 
At each occurrence of new class data or a considerable change in the replay data \(R_t\), the RM network is retrained based on \(R_t\). From Fig.~\ref{fig:embeddings visualization}(c), we can see that the embeddings $E_{tl}$ become linearly separable again after passing through the RM network.


\begin{algorithm}[htbp]
\DontPrintSemicolon
\KwData{base training data \(D_0 = (X_0, Y_0)\), streaming data \(D_t = (X_t, Y_t)\), replay data \(R\) (support set \(S\) and query set \(Q\)), Feature Extractor \(M_f\), Relation Module \(M_r\) }
 \textbf{In the pre-deployment stage:} \;
 Train the model \(M_f\) with \(D_0\); \;
 Save the pre-trained model \(M_f\);\;
 Initiate the replay data \(R_0\) with random samples from the embeddings of \(D_0\) \;
 \textbf{In the streaming stage:} \;
 \While{receiving batch of streaming data \(D_t\)}{
  \(D_t\) = \(D_{tl} \bigcup D_{tu}\) (labelled and unlabelled) \;
  \(E_{tl}\) = \(M_f(D_{tl})\), \(E_{tu}\) = \(M_f(D_{tu})\) \;
  Periodically update \(R_{t}\) based on \(E_{tl}\) \;
  \If{new class or considerable change (\(\frac{1}{4}\) examples of one class have been updated) in \(R_t\)}{
   Update \(R_{t}\) based on \(E_{tl}\) \;
   Randomly select \(S_t\) and \(Q_t\) based on \(R_{t}\) \;
   Retrain \(M_r^t\) \;
   }
  \(E_{tu}\) goes through model \(M_r^t\) to get the classification results \;
 }
 \caption{Workflow of PTRN-HAR}
 \label{algorithm:algorithm 1}
\end{algorithm}

The design principles of PTRN-HAR aimed at achieving higher performance and efficiency are summarized below.
\begin{itemize}
    \item PTRN-HAR prevents catastrophic forgetting by freezing the pre-trained feature extractor which preserves its excellent feature extraction performance for the base classes. Meanwhile, the RM network can reclassify the new class data, enabling the model to learn new tasks. In this way, the model effectively achieves continual learning.
    \item Only the relation module instead of the entire model needs to be retrained during the streaming stage, significantly reducing the computational cost.
    \item PTRN-HAR only needs to store embeddings as replay data, unlike most previous CL methods for HAR \cite{adaimiLifelongAdaptiveMachine2022, jhaContinualLearningSensorbased2021, schiemerOnlineContinualLearning2023} that require storing raw data. This approach greatly reduces the memory needed for storing replay data.
    \item The RM network classifies the embeddings by measuring the similarity between the query embedding and the set of support embeddings. This method not only reduces the amount of data required to train the model but also improves performance compared to previous pre-training and fine-tuning methods that use a multi-layer perceptron or a dense layer as the classifier \cite{maiOnlineContinualLearning2022, cossu2024continual}.
\end{itemize}

\subsection{Data Pre-processing}
\label{sec:Data Pre-Processing}

The raw sensor data undergoes pre-processing through three steps, including data cleansing, synthetic minority oversampling, and data augmentation, before being used to train the feature extractor network.

\textbf{Data cleansing:} We begin by applying linear interpolation to fill NaN values in the raw sensor data. This is followed by segmentation into sliding windows with a fixed window size, and finally normalisation to zero mean and unit variance. 

\textbf{Synthetic minority oversampling:} In HAR scenarios, class imbalance is a common challenge \cite{chen2021deep, guo2021evolutionary}, where certain activity classes are significantly underrepresented compared to the others. This class imbalance can negatively impact the ability of FE to learn generalized representations and can also hinder the effective construction of positive-negative pairs for contrastive learning. To mitigate this issue, we adopt the Synthetic Minority Oversampling Technique (SMOTE) \cite{fernandez2018smote}, which addresses inter-class imbalance by generating synthetic samples for minority classes based on the original dataset. The core idea of SMOTE is to generate synthetic samples for minority classes by interpolating between a sample and one of its nearest neighbors, as shown below:
\[
{x}_{1g} = {x}_1 + \delta \cdot ({x}_{1n} - {x}_1)
\]
where \({x}_{1}\) is a sample from the minority class, \({x}_{1n}\) is one of the k-nearest neighbors of \({x}_{1}\), \(\delta\) is a random number drawn from a uniform distribution in the range \((0,1]\), and \({x}_{1g}\) is the generated sample from \({x}_{1}\). Previous studies \cite{alani2020classifying, guo2019improved} have demonstrated that SMOTE can enhance both classification performance and feature representation quality in imbalanced learning settings. In the context of PTRN-HAR, when data $D_0$ are imbalanced, we apply SMOTE to augment the minority classes. The resulting balanced data are then used as the input for data augmentation to ensure class-wise consistency.


\textbf{Data augmentation:} Given that the primary objective of the FE network in PTRN-HAR is to learn generalizable representations across various activity classes, we employ contrastive loss—a widely adopted method in representation learning—to encourage the extraction of meaningful representations \cite{sakaiSemisupervisedClassificationBased2017, 9873966, wang2021understanding}. Following standard practices in contrastive learning, we apply data augmentation techniques to construct positive and negative pairs \cite{khosla2020supervised, chuang2020debiased}. Specifically, since the input to the PTRN-HAR model is time series data, we adopt four augmentation methods commonly used for sequential signals: jittering, scaling, magnitude warping, and time warping \cite{adaimiLifelongAdaptiveMachine2022, iglesias2023data}. 

Jitter applies additive Gaussian noise \(\epsilon\) to the data, with each channel of the input \(x\) perturbed independently as shown in Equation \ref{jitter}.

\begin{equation}
\label{jitter}
\tilde{x_j} = x + \epsilon, \quad \epsilon \sim \mathcal{N}(0, \sigma^2)
\end{equation}

Scaling multiplies each channel of the input data \(x\) by a random factor \(\alpha\) drawn from a Gaussian distribution:

\begin{equation}
\label{scaling}
\tilde{x}_{i,t,c}^{s} = x_{i,t,c} \cdot \alpha_{i,c}, \quad \alpha_{i,c} \sim \mathcal{N}(1, \sigma^2),
\end{equation}

where \(i, t, c\) represent the sample index, time index, and channel index, respectively. 

Magnitude warping smoothly warps the magnitude of input \(x\) over time using random smooth curves \(s_{i,t,c}\) (Cubic Spline interpolation \cite{um2017data}), as described in Equation \ref{magnitude warping} and \ref{magnitude warping curve}.

\begin{equation}
\label{magnitude warping}
\tilde{x}_{i,t,c}^{m} = x_{i,t,c} \cdot s_{i,t,c}
\end{equation}
\begin{equation}
\label{magnitude warping curve}
s_{i,t,c} = \text{Cubic Spline}(u, r_{i,:,c})(t), \quad r_{i,:,c} \sim \mathcal{N}(1, \sigma^2),
\end{equation}

where \(i, t, c\) represent the sample index, time index, and channel index, respectively.  \(s_{i,t,c}\) is the smooth, time-dependent scaling factor to be applied to the original signal at time \(t\), for sample \(i\),channel \(c\). For equation~\eqref{magnitude warping curve}, \(r_{i,:,c}\) is the vector of random scaling values (the knot values) sampled from a Gaussian distribution, and \(u\) represents fixed vector of time points (the knot positions) evenly spaced along the time axis. Based on \(r_{i,:,c}\) and \(u\) we construct the smooth spline curve \(s_{i,t,c}\) passing through the knot points, which is used as the warp factor for magnitude warping. Similarly, as defined in Equations \ref{time warping} and \ref{time warping curve}, time warping distorts the timeline of input \(x\) (stretches or compresses time intervals) via a smooth warping function, also based on a Cubic Spline.

\begin{equation}
\label{time warping}
\tilde{x}_{i,t,c}^{T} = \text{interp}\left(t, \tau_{i,t,c}, x_{i,\tau_{i,t,c},c} \right)
\end{equation}
\begin{equation}
\label{time warping curve}
\tau_{i,t,c} = \text{Cubic Spline}(u, u \cdot \delta_{i,:,c})(t), \quad \delta_{i,:,c} \sim \mathcal{N}(1, \sigma^2),
\end{equation}

where \(t \in \{0, 1, \dots, T-1\}\) represents the original, uniformly sampled time steps. For each sample \(i\) and channel \(c\), we first generate a nonlinear warping function \(\tau_{i,t,c}\). Then we sample the original signal \(x_{i,:,c}\) at those warped time positions, and finally interpolate the warped signal back onto the original time steps \(t\). In Figure~\ref{fig:pipeline of PTRN-HAR}, we provide a straightforward visualization of these augmentation methods to demonstrate their effectiveness. They are designed to preserve the semantic content of the signals while introducing sufficient variability for contrastive training
\cite{khosla2020supervised}.

\subsection{Feature extractor}
\label{sec:feature extractor}
The FE network in PTRN-HAR is not restricted to any specific architecture; any network capable of effectively extracting features from raw sensor data can be utilized. To balance performance and resource efficiency, PTRN-HAR adopts an LSTM-CNN-based FE network \cite{xiaLSTMCNNArchitectureHuman2020}, as illustrated in Fig.~\ref{fig:pipeline of PTRN-HAR}.


The training of the FE network utilizes two loss functions: cross-entropy loss and contrastive loss. The cross-entropy loss is given in Equation \ref{CrossEntropyLoss}.

\begin{equation}
\label{CrossEntropyLoss}
L_{ce} = -\frac{1}{N} \sum_{n=0}^{N-1} \sum_{c=0}^{C-1} y_{n,c} \log(p_{n,c})
\end{equation}

The core idea of contrastive loss is to cluster similar instances closely while separating dissimilar ones, thereby enhancing the model's ability to extract key features that distinguish between different classes \cite{wang2021understanding}. Given that the pre-deployment stage constitutes a supervised learning scenario, we consider utilizing Supervised Contrastive Learning (SupCon) \cite{khosla2020supervised} to construct more numerous and balanced positive-negative pairs based on label information, thereby fully leveraging the available label information to enhance the model's feature extraction capability. 

For the base training data \(\{x_i, y_i\}_{i=1}^{N}\), since we used 4 augmentation methods for the original data, the augmented dataset are represented as \(\{x_k, y_k\}_{k=1}^{4N}\). 

An anchor \(a \in I \equiv \{1 \ldots 4N\}\) is the index of an arbitrary sample in the augmented dataset \(\{x_k, y_k\}_{k=1}^{4N}\), and \(j(a)\) are the index of all the data in \(\{x_k, y_k\}_{k=1}^{4N}\) labelled with \(y_a\), which is the label for \(x_a\). \(x_a\) and \(x_{j(a)}\) form positive pairs. 

The rest \(\{x_n\mid n \in (A(a) \setminus j(a))\}\) with \(x_a\) form negative pairs, \(A(a) \equiv I \setminus a\).  The contrastive loss \(L_{con}\) of the FE network is calculated by:

\begin{equation}
\label{Contrastive loss}
L_{con}  = \sum_{a \in I} \frac{-1}{|x_{j(a)}|} \sum_{p \in x_{j(a)}} \ln \frac{\exp(E_a \cdot E_p / \tau)}{\sum_{j \in A(a)} \exp(E_a \cdot E_j / \tau)}
\end{equation}

where \(E\) is the embedding features of data, \(\tau\) is the temperature parameter. We use a dot product to measure the similarity between the features of anchor \(E_a\) and positives \(\{E_p \mid p \in x_{j(a)}\}\) to control contrastive loss, which effectively reduces the variance of intra-class features while increasing separability between classes. Moreover, for a given anchor \(a\), the loss function considers the collective influence of all corresponding positive samples by \(\frac{-1}{|x_{j(a)}|} \sum_{p \in x_{j(a)}}\), in contrast to traditional contrastive loss formulations \cite{chuang2020debiased}, which typically construct positive-negative pairs by augmentation on only a single anchor. Using label information more extensively, this approach improves the stability and robustness of feature extraction for HAR. Since this contrastive loss is only calculated based on the augmented data, the final FE's loss function \(L_F\) is the sum of the aforementioned loss functions: \(L_F = L_{ce} + L_{con}\). 


\subsection{Relation Module}
\label{sec:relation module}

Although the FE network is designed to capture generalizable representations, the embeddings of new class data extracted by the frozen FE network during the streaming stage still become linearly inseparable from those of the base classes, as shown in Fig.~\ref{fig:embeddings visualization}(b). Under such conditions, those previous methods, such as the prototypical network \cite{lim2024ssl, Pan_2019_CVPR}, which relies on computing the Euclidean distance between embeddings and class prototypes for classification, and LAPNet-HAR \cite{adaimiLifelongAdaptiveMachine2022}, which employs a dense layer as the classifier, struggle to extract meaningful information from linearly inseparable data and therefore fail to perform effectively. To address this limitation, we proposed a relation module network as the classifier, which substantially enhances the performance of PTRN-HAR.

\subsubsection{Network architecture} 
As shown in Fig.~\ref{fig:pipeline of PTRN-HAR}, the data used to train the RM network are from the replay data \(R\), which contains \(N\) embedding samples (default \(N\) = 20) for each class. The replay data \(R\) are divided into a support set \(S\) and a query set \(Q\). The support set \(S\) is responsible for providing the model with benchmark information for the corresponding class, while the query set \(Q\) is responsible for tuning the model. The RM network uses the knowledge learnt from the support set to correctly classify the samples in the query set. 

Similarly to the core idea of the Relation Network \cite{sungLearningCompareRelation2018}, the support (128 dimensions) and query (128 dimensions) sets are concatenated to obtain a 256-dimensional embedding as input for the convolutional layer. The model classify the query data by finding the most similar one from the support set.
This structure can be regarded as a learnable non-linear comparator, which analyses the similarity between the embeddings by extracting their features. Compared to dense layers or Euclidean distance-based methods \cite{lim2024ssl, Pan_2019_CVPR, adaimiLifelongAdaptiveMachine2022}, the RM Network can re-extract potentially useful latent features from embeddings by comparing the support and query sets. In addition, it fully leverages the information from each sample in the support set, leading to a significant performance improvement. This is evaluated in Section \ref{sec:Ablation study}. 

To avoid overfitting, \(N_s\) (by default, \(N_s\) = 5 in PTRN-HAR) samples of each class are selected from the replay data to form the support set, and the rest as the query set. The model repeats this random selection process in each epoch to make the model utilize the replay data more efficiently.  When the amount of available data is limited, the support and query sets train the network through various random combinations, allowing for more efficient data utilization and significantly improving performance. During inference, PTRN-HAR uses all the replay data as the support set, with the streaming data \(D_t\) serving as the query data. 


\subsubsection{Loss function for RM network}

The output of RM network is the relation score \(r_{ij}\) (given in Equation \ref{eq:relation_score}), which measures the similarity between the embedding of query data \(E_j\) and that of each class in the support set \(E_i\). \(g_{\phi}\) denotes the RM network:

\begin{equation}
r_{ij} = g_{\phi}\left(concatenate \left(E_i, E_j\right)\right), \quad i = 1, 2, \ldots, C
\label{eq:relation_score}
\end{equation}

We use the mean square error (MSE) as the loss function to train the RM network based on the relation score. Similar embeddings have scores closer to 1, and dissimilar ones closer to 0. Additionally, with only a small amount of samples per class (e.g., 20 per class) per class for training, we applied  \(L2\)  regularization to the RM network to avoid overfitting, as defined in Equation (10).

\begin{equation}
\label{loss of relation module}
L_{\text{Relation}} = \sum_{i=1}^{m} \sum_{j=1}^{n} \left( r_{i,j} - 1(y_i == y_j) \right)^2 + \frac{\lambda}{2m} \sum_{i=1}^{k} \theta_i^2
\end{equation}

The first term in Equation~\eqref{loss of relation module} is the MSE loss. \(\lambda\) is the regularization parameter (adjustable), while \(\sum_{i=1}^{k} \theta_i^2\) denotes the sum of squares of the weight parameters (\(L2\) regularization).

\subsubsection{Replay data update}
\label{sec:replay update}
In order to keep the model able to adapt to changes in streaming data and enrich the diversity of the replay data, the replay data also needs to be constantly updated in the streaming stage. In this paper, we refer to the method proposed by Leite and Xiao \cite{leiteResourceEfficientContinualLearning2022} for updating the replay data. As shown in Fig.~\ref{fig:pipeline of PTRN-HAR}, streaming data first goes through the FE network to get the corresponding embeddings. After that, the replay data are updated based on the embeddings according to Equation~\eqref{update replay data},

\begin{equation}
\label{update replay data}
d_{\mathcal{M}} = \sum_k \min_{h \ne k} |t_k - t_h|,
\end{equation}

where \(t\) represents the timestamp of the data. \(d_M\) calculates the sum of the distances between each data sample and its nearest neighbor (in time) data sample. We keep track of the timestamps of all samples in the replay data and regularly update them to maximize \(d_M\), to make the distribution of the replay data more sparse. When the replay data change significantly (i.e., a quarter of the examples of at least one class have been updated), or when a new class appears and needs to be classified, we use the replay data to retrain the RM network. In this way, the model can achieve CL to adapt to the style variation in the streaming data and the appearance of new classes.


\section{Experimental Settings}
\label{sec:settings}

In this section, we first introduce the two different evaluation scenarios used to test the performance of PTRN-HAR, along with the datasets and the evaluation metrics for both performance and resource consumption(Section \ref{sec:datasets}). We then present the baseline methods used for comparison(Section \ref{sec:baselines}). Finally, we describe the implementation details of PTRN-HAR and the baselines, as well as the overall experimental setup.(Section \ref{sec: details})

\subsection{Dataset segmentation and evaluation metrics}
\label{sec:datasets}
\begin{figure}
    \centering
    \includegraphics[width=0.9\linewidth]{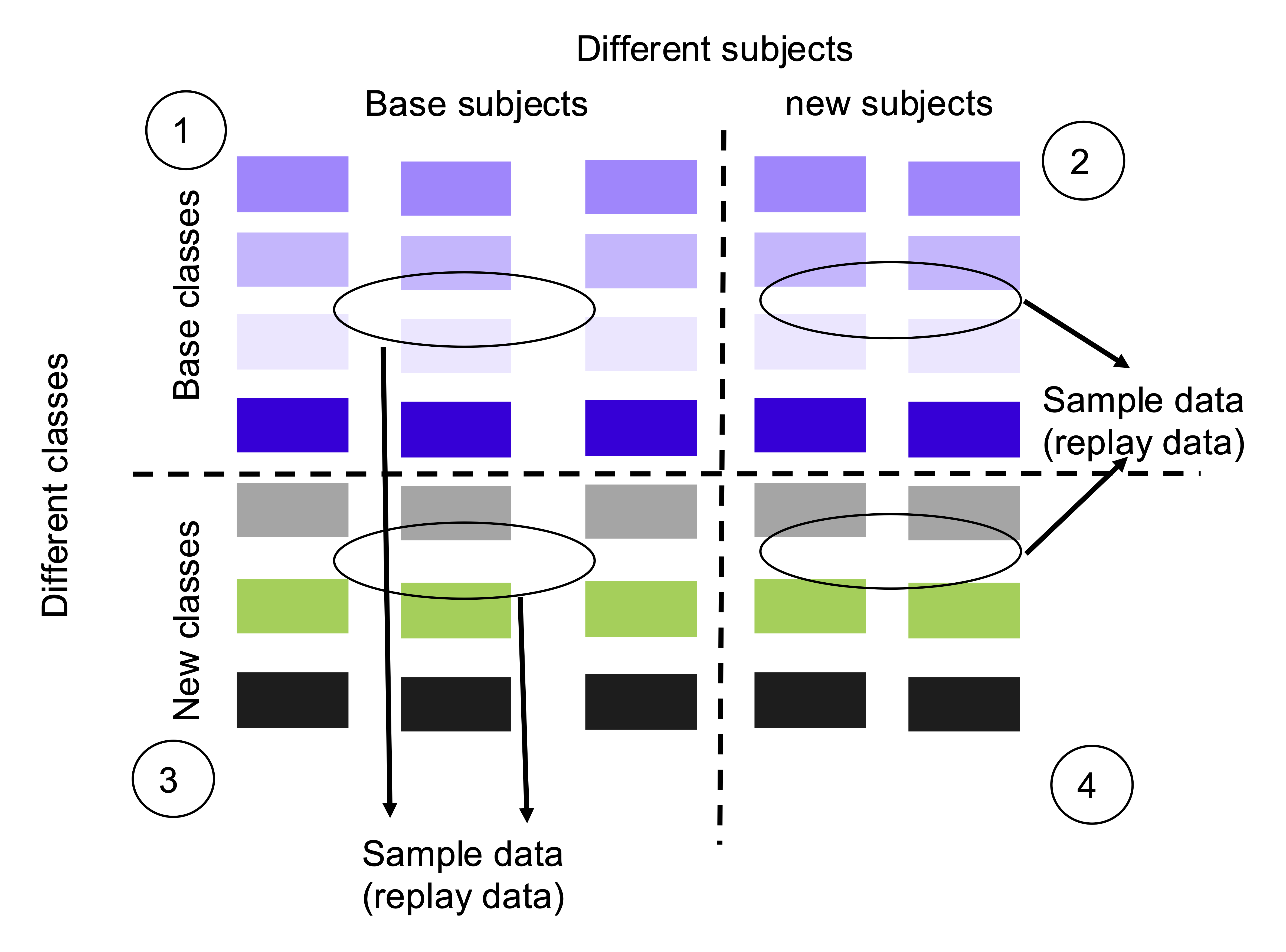}
    \caption{Dataset segmentation based on different OCL scenarios.}
    \label{fig:dataset split}
\end{figure}
PTRN-HAR is tested in OCL scenarios, categorized into within-subject and between-subject settings. As illustrated in Fig.~\ref{fig:dataset split}, we divide the experimental data into 4 groups, based on whether they represent base classes or new classes as well as whether it is collected from the new subjects post-deployment. 

In the within-subject OCL scenario, data from region 1 is used to train the FE network during the pre-deployment stage. We sample \(N\) labelled data per class (default \(N\) = 20) from regions 2 and 4 to train the RM network, and use all data from regions 2 and 4 to test the model. This setup mirrors the real-world scenario, where initial models are trained with data from a different set of subjects and are updated with data from the new subjects -- the end user of the HAR system. We refer to this scenario as within-subject OCL, since both the training and testing data for the RM network during the streaming stage originate from the same subjects. In this case, we assume that new classes of activities may emerge, while the user's ways of performing the existing classes remain consistent. If the same subject's style of performing activities change over time, it can be treated in the same way as between-subject OCL scenarios.

In the between-subject OCL scenario, the ways in which the user performs activities change. Data from region 1 is still used to train the FE network during the pre-deployment stage. However, after deployment, the RM network is trained with data sampled from regions 1 and 3, while the test data come from regions 2 and 4. Utilizing different subjects for training and testing the RM network emulates the real-world scenario where, in addition to the emergence of new classes, the data distribution for the same classes also shifts.

We evaluate PTRN-HAR and baseline methods (Sec. \ref{sec:baselines}) on three public datasets: PAMAP2 \cite{reissIntroducingNewBenchmarked2012}, HAPT \cite{reyes-ortizHumanActivityRecognition2014}, and DSADS \cite{altunComparativeStudyClassifying2010}. A brief overview of each dataset is presented below. The base and new classes are chosen randomly from all three datasets, respectively.

The PAMAP2 dataset \cite{reissIntroducingNewBenchmarked2012} includes 18 activities using three IMUs (placed on the chest, right hand, and left ankle) alongside a heart rate monitor (52 sensor channels). Following previous works \cite{guanEnsemblesDeepLSTM2017, leiteResourceEfficientContinualLearning2022}, to avoid heavy imbalance during the training, 6 of the activities are discarded.  We split the dataset by including the data from participants 5 and 6 in the regions 2 and 4, depending on whether the data belong to the base classes or new classes in Fig.~\ref{fig:dataset split} (new subjects). The remaining data form regions 1 and 3 (base subjects).  For data pre-processing, we segmented the PAMAP2 dataset using a sliding window of \(5.12\) seconds with 78\% overlap, consistent with previous work \cite{leiteResourceEfficientContinualLearning2022, adaimiLifelongAdaptiveMachine2022}.

The HAPT dataset \cite{reyes-ortizHumanActivityRecognition2014} consists of data from 30 subjects performing 12 daily activities. The data was recorded using a smartphone with a single IMU sensor worn on the waist (six sensor channels representing 3D accelerometer and gyroscope readings). A sliding window of \(2.56\) seconds with 50\% overlap was used to segment the data \cite{adaimiLifelongAdaptiveMachine2022, schiemerOnlineContinualLearning2023}. Subjects 29 and 30 are treated as new subjects.

The DSADS dataset \cite{altunComparativeStudyClassifying2010} comprises motion sensor data of 19 daily and sports activities performed by 8 subjects. Each activity was captured for five minutes at 25 Hz. The five IMUs (45 sensor channels) were placed on the torso, right arm, left arm, right leg, and left leg, respectively. Subjects 7 and 8 are treated as new subjects \cite{adaimiLifelongAdaptiveMachine2022, schiemerOnlineContinualLearning2023}.

\begin{table}[htbp]
    \caption{An overview of PAMAP2, HAPT and DSADS}
    \centering
    \resizebox{\linewidth}{!}{
    \begin{tabular}{cccccc}
    \hline
    Datasets & Activity Type & Subjects & \makecell[c]{Channels} & Classes & Balanced \\ 
    \hline
    PAMAP2 & \makecell[c]{Physical Activities} & 9 & 52 & 12 & no \\ 
    \hline
    HAPT & \makecell[c]{Daily Activities \&\\ Postural Transitions} & 30 & 6 & 12 & no \\ 
    \hline
    DSADS & \makecell[c]{Daily \& \\Sports Activities} & 8 & 45 & 19 & yes \\
    \hline
    \end{tabular}
     }
    \label{tab:dataset info}
\end{table}
   
While the DSADS dataset presents balanced classes, PAMAP2 and HAPT exhibit class imbalance. This provides a more comprehensive evaluation of PTRN-HAR's performance across different class distribution scenarios. Given the data imbalance, we use not only accuracy but also the F1-score as a performance metric. To assess the feasibility of deploying PTRN-HAR on edge devices, we evaluate its resources consumption on a Raspberry Pi 5 in terms of FLOPs, training and inference time, CPU and RAM usages, and the size of the replay data.

\subsection{Baselines}
\label{sec:baselines}

In this paper, we implemented in total 8 solutions, including iCaRL(CVPR 2017) \cite{rebuffiIcarlIncrementalClassifier2017}, ER-ACE(ICLR 2022) \cite{caccia2021new}, AGEM(ICLR 2019) \cite{chaudhry2018efficient}, EWC (PNAS 2017) \cite{kirkpatrickOvercomingCatastrophicForgetting2017}, GDumb(ECCV 2020) \cite{prabhu2020greedy},  LAPNet-HAR (Sensors 2022) \cite{adaimiLifelongAdaptiveMachine2022}, joint training and PTRN-HAR. Here, joint training represents the upper-bound performance for CL. Note that joint training does not consist of a CL scenario as it directly trains the model by merging all data. We selected two representatives from rehearsal-based methods (iCaRL \cite{rebuffiIcarlIncrementalClassifier2017}, ER-ACE\cite{caccia2021new}), and regularization-based methods (EWC \cite{kirkpatrickOvercomingCatastrophicForgetting2017}, AGEM \cite{chaudhry2018efficient}), respectively. 

iCaRL and ER-ACE are known for their outstanding performance in rehearsal-based continual learning methods. iCaRL \cite{rebuffiIcarlIncrementalClassifier2017} stores a subset of the most representative data and trains a new model based on the replay data and the new data. ER-ACE \cite{caccia2021new} uses an asymmetric update rule to reduce the overlap between the representations of the new classes and the base classes, and also to reduce the abrupt representation change for all classes.

AGEM and EWC employ regularization-based CL. Before updating model parameters on a new task, AGEM projects the current gradient so that it doesn't conflict with the average gradient of a small memory buffer of past task examples. EWC employs the Fisher Information Matrix (FIM) to preserve important parameters for old tasks.

We also evaluated three recent high-performance rehearsal-based methods in our scenario: CoPE(ICCV 2021) \cite{De_Lange_2021_ICCV}, DER++(NIPS 2020) \cite{buzzega2020dark}, PCR(CVPR 2023) \cite{lin2023pcr}. However, since none of them exceeded the performance of ER-ACE, they were not included as baselines in our study. We did not choose dynamical architecture-based approachs such as PNN \cite{rusuProgressiveNeuralNetworks2022} because their models grow linearly as new classes appear. 

We consider GDumb \cite{prabhu2020greedy} as one of the baselines, as it achieves competitive or even superior performance compared to many CL algorithms on various datasets using a simple greedy strategy - storing the most recent data and retraining model from scratch. It is currently regarded as a strong and effective baseline for evaluating whether a CL algorithm provides truly performance improvement \cite{wang2024comprehensive, yu2023dataset}. We also choose LAPNet-HAR \cite{adaimiLifelongAdaptiveMachine2022}, because it adopts a 'pretraining-then-finetuning' paradigm similar to PTRN-HAR and also takes data efficiency into account. 


\subsection{Implementation details}
\label{sec: details}
Our framework was developed and tested on a system with an Intel Core i5-12600K, 32GB DDR5 RAM, and an NVIDIA RTX A2000 (12GB). Models were implemented in Python 3.10.10 and PyTorch 2.1.0. Since HAR models are often deployed on resource-constrained edge devices, we conducted an additional resource consumption analysis of PTRN-HAR on a Raspberry Pi 5 equipped with a 64-bit Cortex-A76 processor, 8 GB of RAM, and a 16 GB SD card for system installation and environment configuration. We compared PTRN-HAR with other baselines to evaluate it in terms of resource consumption, as summarized in Table~\ref{tab:resource consumption}, to assess its feasibility for real-world deployment. We did not include joint training in Table Table~\ref{tab:resource consumption}, as it involves training on data from all classes using the FE network in a non-incremental manner to find the upper bound for continual learning performance. Therefore, it is not directly comparable to other methods in terms of resource consumption.

In the pre-deployment stage, we used grid search with five-fold cross-validation to optimize the hyperparameters for the FE network. In the streaming stage, to ensure resource efficiency and handle limited data, we avoided grid search for the RM network. Overfitting was alleviated by randomly selecting the support and query sets and applying \(L2\) regularization. For the RM network, we used a batch size of 50, 50 epochs, and a replay size of 15 samples per class for HAPT, and 20 samples per class for PAMAP2 and DSADS. Due to the limited number of samples for certain classes in the HAPT test set, we reduced the replay data size to 15 for HAPT to ensure that the replay data contains only a small portion of the test data. We employed Adam optimizer with learning rate=1e-3. All baselines were implemented using the Avalanche framework \cite{lomonaco2021avalanche} and their hyperparameters were optimized via grid search. To ensure a fair comparison, all baselines used the same feature extractor architecture (LSTM-CNN) as used in PTRN-HAR.

Since PTRN-HAR incorporates data augmentation when constructing the contrastive loss, we conducted a comparative analysis to evaluate the impact of data augmentation across all baselines. Most methods exhibited a performance improvement of \(2\text{-}4\%\), and none showed any degradation. Consequently, all baseline performance results reported in Table \ref{tab:comparison of Accuracy and Macro F1 in OCL scenarios} and Fig.~\ref{fig:different number of new class} are based on models trained with augmented data. However, for the resources consumption evaluation (Table~\ref{tab:resource consumption}), all baselines were tested using the original, non-augmented data to ensure a fair comparison by avoiding additional computational overhead introduced by data augmentation.

\section{Results}
\label{sec:results}
Different base classes can influence the feature extractor's ability, which may subsequently impact the model's performance. The results in this section are based on experiments conducted with five different base class combinations, randomly selected, and averaged. First, we compare the performance and resource consumption of PTRN-HAR with baselines (Section \ref{sec:Continual learning score} and \ref{sec:Resource efficiency}). We then present how the model's performance changes when varying the number of base classes (Section \ref{sec:Impact of the Number of base class}) and the amount of labelled data available for each new class (Section \ref{sec:Impact of available labeled data per new class}). Finally, we conduct the ablation study to assess the contribution of key designs of the model (Section \ref{sec:Ablation study}). The number of base classes for all the above experiments was set to 7 for HAPT and PAMAP2, and 10 for DSADS, with the remaining classes used as new classes (5 for HAPT and PAMAP2 and 9 for DSADS). In Section \ref{sec:Impact of the Number of base class}, when the number of base classes varies, the number of new classes is fixed to 5.

\subsection{Performance}
\label{sec:Continual learning score}

\subsubsection{Overall performance}
The results of the performance comparison between PTRN-HAR and the baseline methods under different OCL scenarios are summarized in Table \ref{tab:comparison of Accuracy and Macro F1 in OCL scenarios}. In the within-subject scenario, PTRN-HAR outperforms all baselines across all three datasets in terms of both accuracy and Macro-F1 score. Specifically, PTRN-HAR achieves approximately \(3\text{–}5\%\) higher accuracy and demonstrates a substantial improvement of \(16.7\text{–}18.9\%\) in Macro-F1 score compared to the baselines. Macro-F1 score provides a fair assessment of a model's performance across all classes and is particularly important in HAR applications, where class imbalance is often severe \cite{alani2020classifying, meyer2022u}. PTRN-HAR's superiority in terms of Macro-F1 stems from two key design choices. First, the use of contrastive loss enables the construction of more balanced positive and negative pairs across classes, leading to more stable and generalizable feature representations. Second, the RM network selects an equal number of representative samples per class as replay data, effectively mitigating the negative impact of class imbalance on model performance.

\begin{table*}[htbp]
    \caption{Comparison of performance in terms of accuracy and macro-F1 score for PTRN-HAR and baselines in OCL scenarios.}
    \centering
    \resizebox{\linewidth}{!}{
    \begin{tabular}{lcccccccccccc}
        \hline
        Scenario & \multicolumn{6}{c}{Within-subject scenario} & \multicolumn{6}{c}{Between-subject scenario} \\
        Dataset & \multicolumn{2}{c}{PAMAP2} & \multicolumn{2}{c}{HAPT} & \multicolumn{2}{c}{DSADS} & \multicolumn{2}{c}{PAMAP2} & \multicolumn{2}{c}{HAPT} & \multicolumn{2}{c}{DSADS} \\
        Metric (\%) & Accuracy & Macro-F1 & Accuracy & Macro-F1 & Accuracy & Macro-F1 & Accuracy & Macro-F1 & Accuracy & Macro-F1 & Accuracy & Macro-F1 \\ 
        \hline
        Joint\_training & 86.60 & 85.30 & 80.72 & 69.46 & 80.96 & 86.94 & 86.60 & 85.30 & 80.72 & 69.46 & 80.96 & 86.94 \\
        ER-ACE & 79.35 & 63.47 & 82.75 & 53.54 & 82.60 & 64.19 & \textbf{74.96} & 64.09 & 73.88 & 53.47 & 72.71 & 58.88 \\
        ICaRL & 66.91 & 61.68 & 71.97 & 49.48 & 70.57 & 54.44 & 72.88 & 61.24 & 61.55 & 48.63 & 62.03 & 50.23 \\
        GDumb & 73.60 & 51.69 & 60.65 & 51.28 & 74.32 & 56.58 & 70.26 & 50.05 & 52.58 & 42.88 & 57.14 & 50.55 \\
        AGEM & 18.82 & 17.04 & 3.36 & 17.67 & 13.23 & 14.76 & 14.83 & 15.86 & 2.35 & 15.01 & 12.08 & 12.71 \\
        LAPNet-HAR & 71.70 & 55.09 & 28.28 & 21.75 & 45.63 & 33.41 & 65.92 & 51.86 & 32.17 & 24.83 & 43.57 & 36.53 \\
        EWC & 10.37 & 16.28 & 49.44 & 28.21 & 6.25 & 9.93 & 13.80 & 15.60 & 2.35 & 14.94 & 6.25 & 9.63 \\
        \textbf{PTRN-HAR} & \textbf{81.02} & \textbf{80.19} & \textbf{84.30} & \textbf{72.44} & \textbf{85.39} & \textbf{82.89} & 74.56 & \textbf{76.00} & \textbf{82.06} & \textbf{68.06} & \textbf{76.75} & \textbf{74.21} \\
        \hline
    \end{tabular}
    }
    \label{tab:comparison of Accuracy and Macro F1 in OCL scenarios}
\end{table*}

In the results of baselines, it is evident that the regularization-based CL methods: EWC and AGEM perform poorly, aligning with the findings of Jha et al.\cite{jhaContinualLearningSensorbased2021}. Similar results between EWC and AGEM stem from severe catastrophic forgetting, which means that the model is only capable of classifying the classes appeared in the final batch. The catastrophic forgetting observed in EWC and AGEM, which was not reported by the original authors \cite{kirkpatrickOvercomingCatastrophicForgetting2017, liuRotateYourNetworks2018, liLearningForgetting2017}, is due to our challenging testing conditions: under an OCL scenario with limited labelled data for new classes.

In the between-subject scenario, there is a significant decrease in overall performance compared to the within-subject OCL scenario. However, PTRN-HAR still maintains an overall superior performance, with a relative improvement of \(12\text{–}16\%\) in Macro-F1 score compared to ER-ACE. The only exception is a slight drop in accuracy on the PAMAP2 dataset, where PTRN-HAR performs marginally below ER-ACE.


\begin{figure*}
    \centering
    \includegraphics[width=0.95\linewidth]{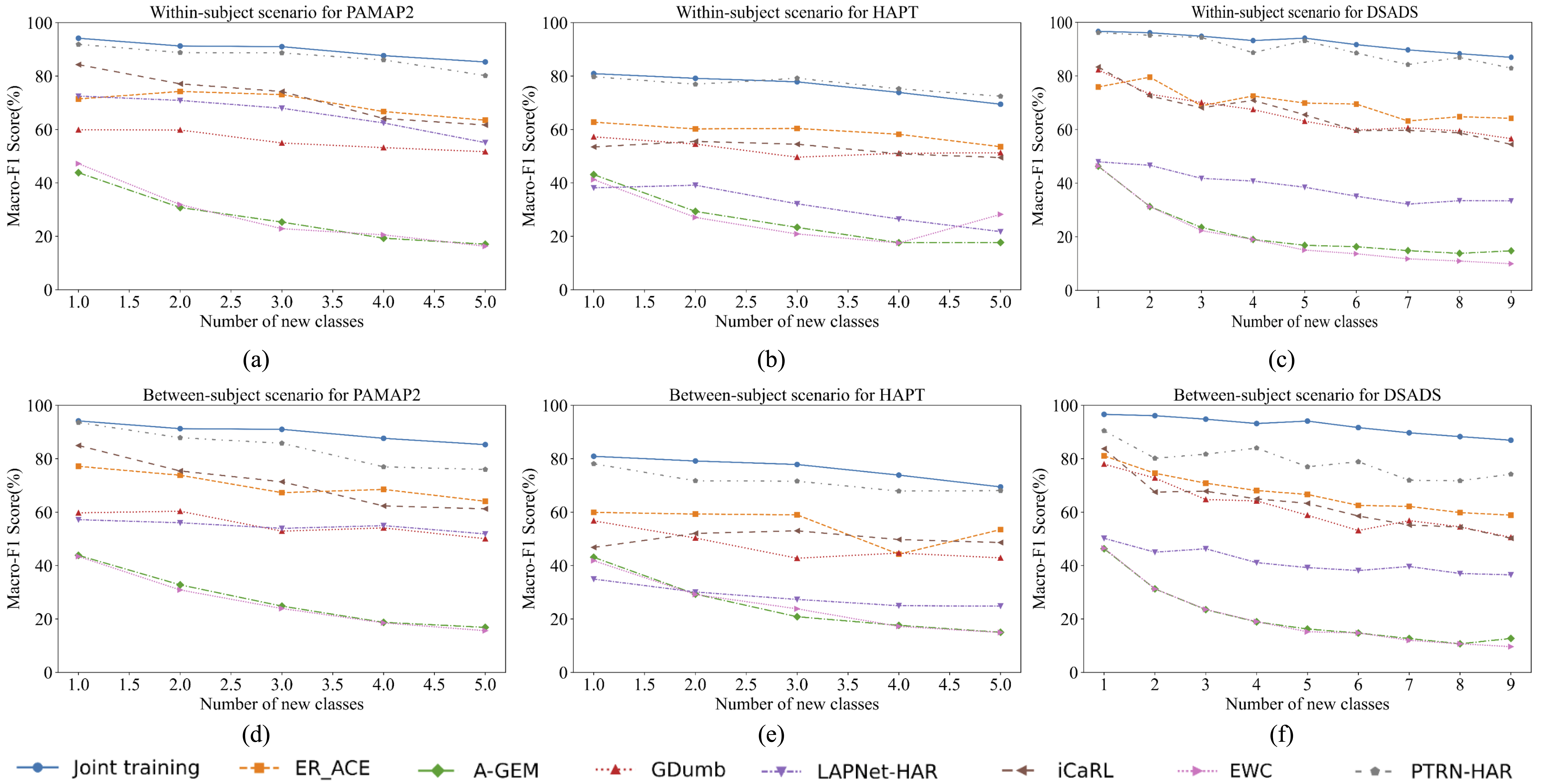}
    \caption{Comparison of Macro F1 score with different number of new classes in different OCL scenarios. }
    \label{fig:different number of new class}
    \vspace{-0.25cm}
\end{figure*}

\subsubsection{Performance decrease with the number of new classes}

We further evaluate PTRN-HAR's performance as the number of new classes incrementally increases, a scenario common in practical CL applications. The streaming stage is divided into multiple subtasks, a new class is added during each subtask. The streaming data contains both old and new classes to meet the requirements of OCL scenarios. Based on this experiment, we can examine the model's stability and determine if the performance degrades significantly as the number of new classes increases. 

As illustrated in Fig.~\ref{fig:different number of new class}, all methods experience a decrease in performance as the number of new classes increases; however, PTRN-HAR exhibits a much smaller decrease compared to the baseline methods. This is because the source of PTRN-HAR's CL ability is different from the other methods, which actually retrain the whole model continuously in the streaming stage to achieve continual learning. Specifically, PTRN-HAR can achieve CL because the frozen FE network trained based on contrastive loss can extract embeddings that contain useful features of the raw sensor data, even if these data belong to an unseen new class. Therefore, as long as the FE network trained on base classes has sufficient generalizability to extract general features, even if the number of new classes keeps increasing, it will not affect the model's feature extraction. Though the RM network has some difficulties in reclassifying these embeddings, which leads to a certain performance decrease, but the magnitude of the decrease is lower than that of other methods.  

Another notable observation is that in Fig.~\ref{fig:different number of new class}(b) and (e), PTRN-HAR even outperforms joint training in certain cases. This phenomenon is particularly evident on the HAPT dataset, which exhibits severe class imbalance due to the inclusion of transitional activities—such as sitting to standing or lying to standing—that have very few corresponding samples. Joint training tends to underperform on these underrepresented classes, leading to a significant drop in Macro-F1 score, as this metric assigns equal importance to all classes. In contrast, PTRN-HAR is more robust under such conditions to handle extreme class imbalance without neglecting rare classes.

\subsection{Resource efficiency}
\label{sec:Resource efficiency}

To evaluate the deployment feasibility and resource consumption of PTRN-HAR on edge devices, we conducted resources consumption evaluation on a Raspberry Pi 5. The results, presented in Table~\ref{tab:resource consumption}, show that the average training time per class during the streaming stage for PTRN-HAR ranges from 264 to 391 seconds, while another high-performing method, ER-ACE, requires between 308 and 592 seconds. Consequently, for a 10-class classification task, complete retraining of the RM network would take approximately 44 to 65 minutes. Given that the model only needs to be updated periodically, such as once a day or once a week, this level of computational cost is acceptable for real-world deployment on resource-constrained devices.

PTRN-HAR requires fewer floating-point operations per second (FLOPS) compared to the baselines, primarily because the input for the RM network consists of low-dimensional embedding pairs (size: 256). This architectural advantage also contributes to significantly shorter training times for PTRN-HAR on most datasets—except for DSADS. In the case of DSADS, the original input data has relatively small dimensions, making the compression from raw data to embeddings less pronounced. As a result, the training time of PTRN-HAR on DSADS is slightly higher than that of ER-ACE. On the other hand, since the FE network in PTRN-HAR remains frozen during the streaming stage, part of the computational cost is shifted to the pre-deployment stage. This design choice allows the model to retain strong feature extraction capabilities without incurring significant computational overhead to train the FE network during the streaming stage.

\begin{table*}[htbp]
\renewcommand{\arraystretch}{0.9}
    \caption{Comparison of resource consumption metrics for different methods. Unavailable results are denoted with ‘-’}
    \centering
    \resizebox{\linewidth}{!}{
    \begin{tabular}{lccccccccc}
        \hline
        Method & \multicolumn{3}{c}{PTRN-HAR} & \multicolumn{3}{c}{LAPNet-HAR}  & \multicolumn{3}{c}{ER\_ACE} \\ 
        Dataset &
        DSADS & HAPT & PAMAP2 &
        DSADS & HAPT & PAMAP2 &
        DSADS & HAPT & PAMAP2 \\ 
        \hline
        FLOPs &
        0.386M & 0.386M & 0.386M &
        25.65M & 25.19M & 116.47M &
        18.06M & 17.86M & 74.45M \\
        Training time/per class (s) &
        391.62 & 263.23 & 264.15 &
        783.48 & 1377.21 & 1404.27 &
        307.88 & 528.61 & 591.76 \\
        Inference time/per sample (ms) &
        7.23+3.69 & 6.18+2.61 & 7.73+2.59 &
        13.49 & 16.90 & 18.65 &
        7.23 & 6.183 & 7.73 \\
        Train/test RAM usage (MB) &
        1712/1889 & 1111/1232 & 2688/2760 &
        3263/2194 & 1738/1227 & 7491/5309 &
        2410/1743 & 1503/850 & 3604/3578 \\
        Train/test CPU usage (\%) &
        385.2/384 & 370.2/393.7 & 383.9/387.5 &
        384.3/369.6 & 375.3/358.0 & 388.9/372.1 &
        384.6/352.9 & 368.8/340.8 & 392.7/383.4 \\
        Replay data size (KB) &
        190 & 120 & 120 &
        1140 & 5273 & 24960 &
        1140 & 5273 & 24960 \\         
        \hline
        \hline
        Method &\multicolumn{3}{c}{iCaRL} & \multicolumn{3}{c}{GDumb} & \multicolumn{3}{c}{AGEM}\\ 
        Dataset &
        DSADS & HAPT & PAMAP2 &
        DSADS & HAPT & PAMAP2 &
        DSADS & HAPT & PAMAP2 \\ 
        \hline
        FLOPs &
        18.06M & 17.86M & 74.45M &
        18.06M & 17.86M & 74.45M &
        18.06M & 17.86M & 74.45M \\
        Training time/per class (s) &
        408.16 & 593.60 & 653.89 &
        359.62 & 451.66 & 465.73 &
        314.92 & 523.84 & 589.09 \\
        Inference time/per sample (ms) &
        7.22 & 6.206 & 7.73 &
        7.23 & 6.183 & 7.73 &
        7.23 & 6.183 & 7.73 \\
        Train/test RAM usage (MB) &
        2914/1702 & 1945/1058 & 4914/2917 &
        2051/2069 & 886/934 & 4176/3937 &
        2382/1784 & 1523/877 & 6206/3427 \\
        Train/test CPU usage (\%) &
        388.5/363.0 & 381.9/377.3 & 392.1/373.6 &
        370.9/368.0 & 355.8/343.5 & 372.2/375.8 &
        381.7/357.3 & 371.3/364.6 & 390.9/380.5 \\
        Replay data size (KB) &
        1140 & 5273 & 24960 &
        1140 & 5273 & 24960 &
        1140 & 5273 & 24960 \\
        \hline
        \hline
        Method &\multicolumn{3}{c}{EWC} & \multicolumn{3}{c}{ - } & \multicolumn{3}{c}{ - }\\ 
        Dataset &
        DSADS & HAPT & PAMAP2 &
        - & - & - &
        - & - & - \\ 
        \hline
        FLOPs &
        18.06M & 17.86M & 74.45M &
        - & - & - &
        - & - & - \\
        Training time/per class (s) &
        310.21 & 518.96 & 574.73 &
        - & - & - &
        - & - & - \\
        Inference time/per sample (ms) &
        7.73 & 6.183 & 7.73 &
        - & - & - &
        - & - & - \\
        Train/test RAM usage (MB) &
        2450/1785 & 1472/917 & 6007/3426 &
        - & - & - &
        - & - & - \\
        Train/test CPU usage (\%) &
        386.7/357.8 & 377.0/347.4 & 389.2/386.9 &
        - & - & - &
        - & - & - \\
        Replay data size (KB) &
        1140 & 5273 & 24960 &
        - & - & - &
        - & - & - \\
        \hline 
    \end{tabular}
    }
    \label{tab:resource consumption}
\end{table*}

As shown in Table~\ref{tab:resource consumption}, the inference time of PTRN-HAR is slightly higher than that of the baselines due to its two-stage architecture comprising the FE and RM networks. The FE network requires approximately 6 to 7 ms to process a sample and generate its embedding - comparable to the baselines, which share the same feature extractor architecture. The RM network then classifies the embedding, taking an additional 2.6–3.7 ms. As a result, the total inference time of PTRN-HAR is approximately 10 ms per sample, which remains acceptable for real-world deployment.

Table~\ref{tab:resource consumption} presents the usage of RAM and CPU during both training and inference for all methods, which further supports the feasibility of PTRN-HAR for real-world deployment. PTRN-HAR only needs to store the embeddings of the data, while all other replay-based methods must store the raw sensor data. As raw sensor data are approximately 6 to 44 times larger than the embeddings, under identical memory constraints, PTRN-HAR can either achieve substantial memory savings or allocate space for storing more samples, thereby enhancing performance.


\subsection{Impact of the number of base classes}
\label{sec:Impact of the Number of base class}
\begin{figure}
    \centering
    \includegraphics[width=0.9\linewidth]{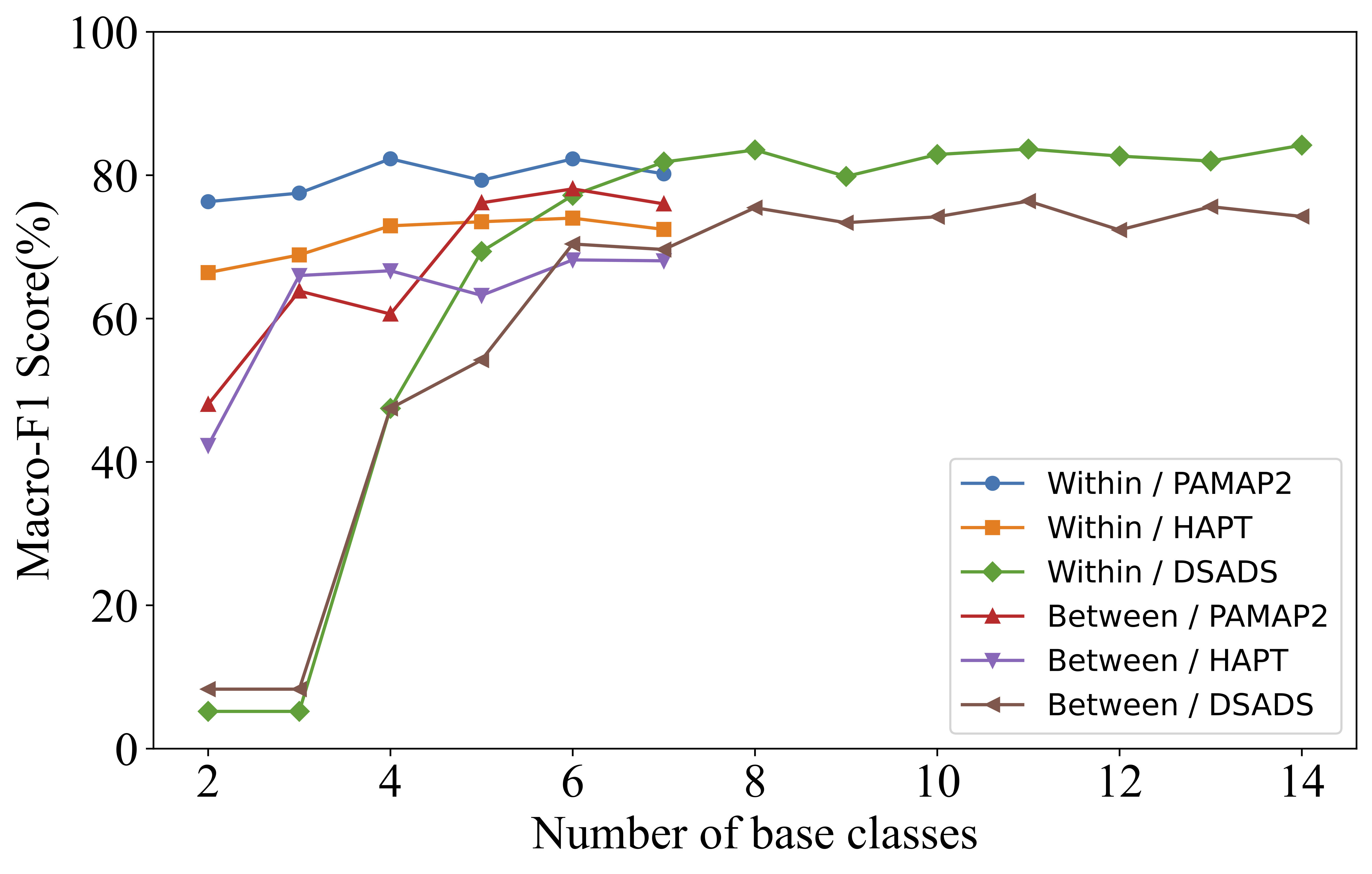}
    \caption{The Macro F1 scores when different number of base classes are used for training FE network.}
    \label{fig:different base class num}
\end{figure}

The results of training the FE network with varying numbers of base classes are presented in Fig.~\ref{fig:different base class num}. It can be observed that when the number of base classes is too small, the model occasionally fails to converge, indicating that the FE network struggles to learn generalizable features with insufficient class diversity. Under the within-subject scenario, no convergence failures are observed on the PAMAP2 and HAPT datasets. However, on the DSADS dataset, the model fails to converge when trained with only 2 or 3 base classes. Once the number of base classes reaches five or more, further increases yield only marginal improvements in accuracy. This suggests that PTRN-HAR can achieve satisfactory performance without requiring a large number of base classes, highlighting its data efficiency in real-world applications.

The between-subject scenario is more challenging, and therefore the model fails to converge in both HAPT and DSADS when the number of base classes is less than 3. As the number of base classes increases, the performance improvement in case of PTRN-HAR is more evident compared to the within-subject scenario. This indicates that, in this scenario, because the training data and test data involve different subjects, additional data are required to enhance the model's generalization and achieve better performance.

\begin{figure}
    \centering
    \includegraphics[width=0.9\linewidth]{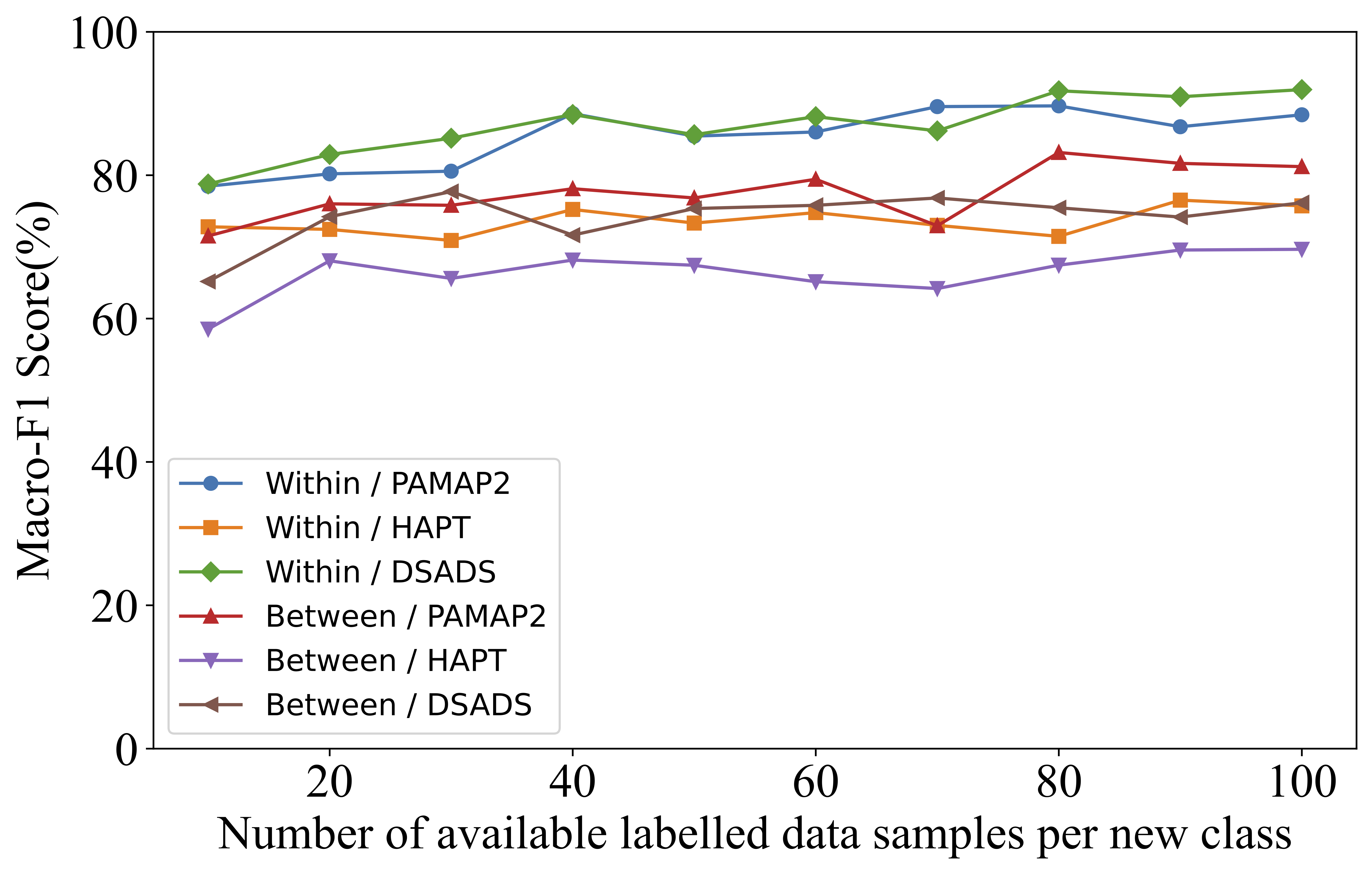}
    \caption{The performance for different availability of labelled samples per new class.}
    \label{fig:different replay data size}
    \vspace{-0.25cm}
\end{figure}

\subsection{Impact of available labelled data per new class}
\label{sec:Impact of available labeled data per new class}
Fig.~\ref{fig:different replay data size} shows the performance of PTRN-HAR with different amounts of labelled data per new class available. The performance of the model shows a slight improvement with an increasing amount of available labelled data, but PTRN-HAR is relatively insensitive to the availability of labelled data per new class. Even with as few as 10 labelled samples per class in the replay data, there is no significant drop in model performance.

\subsection{Ablation study}
\label{sec:Ablation study}

\begin{table*}[tb]
    \caption{Ablation study for PTRN-HAR on DSADS, HAPT, and PAMAP2 under within subject scenario. Metric: Macro F1 Score. DeepConvLSTM \cite{ordonezDeepConvolutionalLstm2016} and LSTM-CNN \cite{xiaLSTMCNNArchitectureHuman2020} are the two different FE network structures tested. '-' for RM means replacing relation module with a three-layer MLP as classifier.}
    \centering
    \resizebox{\linewidth}{!}{
    \begin{tabular}{cccccccccccccc}
        \hline
        \multicolumn{5}{c}{} & \multicolumn{3}{c}{DSADS} & \multicolumn{3}{c}{HAPT} & \multicolumn{3}{c}{PAMAP2} \\ 
         & DeepConvLSTM & LSTM-CNN & RM & \(L_{con}\) & Base class & New class & Overall & Base class & New class & Overall & Base class & New class & Overall \\ 
        \hline
        \#1 & - & \checkmark & \checkmark & - & 53.18 & 71.36 & 60.76 & 72.64 & 66.13 & 69.93 & 73.43 & 64.82 & 69.84 \\ 
        \#2 & - & \checkmark & - & \checkmark & 62.15 & 28.31 & 46.12 & 70.13 & 27.14 & 52.22 & 70.15 & 22.13 & 50.14 \\ 
        \#3 & \checkmark & - & \checkmark & \checkmark & 72.19 & 49.91 & 61.64 & 72.2 & 47.13 & 61.75 & 73.97 & 43.16 & 61.13 \\
        \#4 & - & \checkmark & \checkmark & \checkmark & \textbf{86.71} & \textbf{78.65} & \textbf{82.89} & \textbf{76.19} & \textbf{67.24} & \textbf{72.46} & \textbf{89.28} & \textbf{67.46} & \textbf{80.19} \\ 
        \#5 & \multicolumn{4}{c}{fine-tuning PTRN-HAR} & 87.36 & 87.21 & 87.29 & 85.46 & 73.6 & 80.52 & 92.49 & 75.83 & 85.55 \\
       \hline
    \end{tabular}
    }
    \label{table:ablation study}
\end{table*}

To evaluate the contribution of key designs in PTRN-HAR, we conducted an ablation study focusing on two key elements: the Relation Module (RM) network and contrastive loss \(L_{con}\). The results are summarized in Table~\ref{table:ablation study}.

By comparing Experiments \#4 and \#2, we observed a notable performance drop when replacing the RM network with a three-layer MLP. This suggests that the RM network plays a crucial role, particularly in scenarios with limited data for new classes, where an MLP struggles to effectively learn discriminative features. Furthermore, a performance decline is also observed on the base classes, indicating that the RM network contributes not only to new classes recognition but also enhances the classification of base class data.

The impact of contrastive loss is assessed by comparing Experiments \#4 and \#1. Removing \(L_{con}\) leads to only a slight decrease in Macro-F1 score on new classes, implying that classification of novel classes relies primarily on the RM network. However, for base classes, the absence of contrastive loss during FE network training reduces inter-class separability and increases intra-class variability, resulting in a more significant decline in Macro-F1 score.

In Experiments \#3 and \#4, replacing the original FE network with a DeepConvLSTM architecture \cite{ordonezDeepConvolutionalLstm2016} causes an overall degradation in performance. This confirms the effectiveness of the LSTM-CNN-like structure as the FE network in this study and highlights that the feature extraction capability of the FE network significantly impacts the CL performance of PTRN-HAR.

Since the FE network in PTRN-HAR remains frozen during the streaming phase, experiment \#5 evaluates the model's performance when the FE network is continuously fine-tuned with the streaming data. The results show improvements of approximately 5\%, 8\%, and 5\% in case of DSADS, HAPT, and PAMAP2, respectively. However, this approach comes at the cost of significantly increased training time and memory usage during the streaming stage. Therefore, in this paper, the fine-tuning scheme is not adopted to prioritize resource efficiency.

\section{Conclusion}
\label{sec:conclusion}
This paper presents PTRN-HAR, a resource-efficient OCL approach for sensor-based HAR. PTRN-HAR is the first to use a relation module in CL for HAR, replacing the dense classification layer. This design choice reduces the data required for CL by leveraging the similarity between embeddings. We demonstrate that during CL, training only the relation module, while freezing the feature extractor, significantly reduces computational cost. Additionally, PTRN-HAR incorporates contrastive loss in the pre-deployment stage, decreasing intra-class distance and increasing inter-class distance, thus improving knowledge preservation. Comparative evaluations on 3 public datasets and both within-subject and between-subject scenarios show that PTRN-HAR matches or exceeds the performance of baseline methods while substantially improving data efficiency, reducing computational cost, and memory usage. Ablation studies confirm the effectiveness of both the relation module and contrastive loss.

Although PTRN-HAR shows superior performance compared with the baselines, it can be further improved in the future from the following perspectives. The pretraining of the FE network is based on a subset of the original dataset, which may constrain the generalization ability of the extracted features. A potential future direction is to leverage LLMs to create diverse synthetic data for more robust feature extractor training. Moreover, since PTRN-HAR periodically updates the RM network on edge devices based on replay data while keeping the FE network frozen, integrating federated learning offers another promising enhancement. In such a setup, the RM network could be updated locally on edge devices to preserve personalization and data privacy, while the FE network could be periodically updated in the cloud to improve global feature extraction performance.

 \begingroup
\bibliographystyle{IEEEbib}
\bibliography{ref}
\endgroup

\vfill

\end{document}